\documentclass[lettersize,journal]{IEEEtran}
\usepackage{amsmath,amsfonts}
\usepackage{algorithmic}
\usepackage{array}
\usepackage[caption=false,font=normalsize,labelfont=sf,textfont=sf]{subfig}
\usepackage{textcomp}
\usepackage{stfloats}
\usepackage{url}
\usepackage{verbatim}
\usepackage{graphicx}

\usepackage{array}
\usepackage{booktabs}
\usepackage{makecell}
\usepackage{tabularx}

\newcolumntype{L}[1]{>{\raggedright\arraybackslash}p{#1}}
\newcolumntype{C}[1]{>{\centering\arraybackslash}p{#1}}
\newcolumntype{Y}{>{\raggedright\arraybackslash}X}

\def\BibTeX{{\rm B\kern-.05em{\sc i\kern-.025em b}\kern-.08em
    T\kern-.1667em\lower.7ex\hbox{E}\kern-.125emX}}
\usepackage{balance}
\begin{document}
\title{BranchDrive: A Branch-Structured Dataset for Action-Conditioned Driving Prediction}
\author{Feeza Khan Khanzada, Sudarshan Sridhar and Jaerock Kwon

\thanks{*This work was supported in part by the National Science
Foundation (NSF) under Grant 2500638.}
\thanks{Feeza Khan Khanzada, Sudarshan Sridhar and Jaerock Kwon are with the Department of Electrical and Computer Engineering, University of Michigan-Dearborn, 4901 Evergreen Rd, Dearborn, MI 48128, United States. {\tt\small \{feezakk, ssushi, jrkwon\}@umich.edu}}%
}


\maketitle

\begin{abstract}
Most autonomous-driving datasets record only the action executed by a behavior policy and the single future that followed, providing limited supervision for comparing alternative ego decisions. We introduce BranchDrive, a branch-structured CARLA dataset and benchmark that pairs one canonical pre-decision history with one nominal expert future and twelve physically executed intervention futures spanning acceleration, braking, and left- and right-steering policies at three magnitudes. Each intervention lasts 2.5 s and is followed by expert recovery. Following control-compliance, modality-completeness, replay-fidelity, and action-leakage audits, the frozen benchmark contains 606 independent branch groups and 7,878 associated trajectories. We evaluate prediction of six continuous short-horizon outcomes and a ten-step ego trajectory using action-only, history-only, structured, visual, multimodal, and privileged bird's-eye-view models. On the held-out test split, the structured history-and-action model achieves a macro normalized mean absolute error of 0.5036 and an average displacement error of 2.2042 m, significantly outperforming both restricted baselines. In full-information offline evaluation, its outcome-derived selector increases balanced policy value from 0.5364 to 0.5704 and reduces normalized regret from 0.2674 to 0.1495 relative to the frozen action prior. However, a validation-calibrated minimum-separation guard rejects every intervention, showing that conservative execution remains unresolved. BranchDrive therefore supports action-conditioned short-horizon prediction and fixed-bank offline decision evaluation, but does not establish exact causal effects, binary safety prediction, or closed-loop safety improvement.
\end{abstract}


\section{Introduction}
\label{sec:introduction}

Autonomous-driving systems are ultimately judged by the consequences of their decisions rather than by their ability to reproduce the action of a human or privileged expert. This distinction is especially important in safety-relevant
situations, where several maneuvers may initially appear plausible but produce substantially different futures. When approaching a blocked lane, for example, a vehicle may retain the expert maneuver, brake, steer around the obstruction,
or accelerate through the conflict region. The consequence of each choice depends jointly on the scene geometry, surrounding traffic, ego dynamics, maneuver magnitude, and subsequent controller response. A decision model must therefore reason not only about what is likely to happen, but about what is
likely to happen under each candidate action.

Large-scale driving datasets have enabled substantial progress in perception, motion forecasting, planning, and end-to-end control \cite{caesar2020nuscenes,sun2020waymo,wilson2021argoverse2,caesar2021nuplan}. Most such datasets are observational: for a driving history $H_i$, they record the action $a_i$ executed by the logging policy and the single future $\tau_i$ that followed. Their basic supervision is
therefore
\begin{equation}
    H_i \longrightarrow (a_i,\tau_i),
\end{equation}
while the outcomes of actions not selected by the behavior policy remain unobserved. This structure is appropriate for imitation learning and single-future forecasting, but it does not directly supervise comparisons among alternative ego decisions. An action-conditioned model queried with an
unobserved braking, steering, or acceleration maneuver must extrapolate beyond the action actually taken. This limitation is particularly consequential in rare interactive situations, where the action distribution is imbalanced and an incorrect extrapolation can reverse the preferred decision.

Closed-loop simulation provides a means to execute alternative actions without exposing real road users to risk. Platforms and benchmarks based on CARLA have enabled controlled evaluation of complete policies across interactive routes, hazard scenarios, and system perturbations \cite{dosovitskiy2017carla,xu2022safebench,jia2024bench2drive}. However, their principal data unit is typically a route, scenario, or complete policy rollout. Alternative actions
may be tested by repeatedly running a simulator, but their outcomes are not generally released as a persistent, aligned action bank associated with one decision context. Action-conditioned world models provide another form of
alternative-future reasoning, but their candidate futures are generated by the model itself. Consequently, errors in future synthesis and errors in action evaluation are confounded. A reusable decision dataset instead requires multiple physically executed futures, explicit intervention semantics, and
measured evidence that the corresponding pre-intervention histories are sufficiently comparable.

This paper introduces \emph{BranchDrive}, a branch-structured CARLA dataset and benchmark for action-conditioned driving prediction and offline decision evaluation. The atomic unit is a branch group
\begin{equation}
    G_i =
    \left(
        H_i,
        \left\{
            e_k,\,
            \tau_{i,k},\,
            y_{i,k},\,
            m_{i,k}
        \right\}_{k=0}^{12},
        q_i
    \right),
\end{equation}
where $H_i$ is a canonical pre-decision history, $e_k$ encodes candidate policy $k$, $\tau_{i,k}$ is its simulator-executed future trajectory, $y_{i,k}$ contains its continuous outcomes, $m_{i,k}$ is an availability mask, and $q_i$ contains collection- and replay-quality metadata. Each retained
group contains one nominal expert future and twelve intervention futures: accelerate-straight, braking, left-steering, and right-steering policies, each evaluated at three magnitudes. The candidates are temporally extended policies
with distinct lateral and longitudinal control semantics rather than arbitrary action classes or independent scalar actuator perturbations.

BranchDrive is collected through a two-pass procedure. A privileged expert first executes the complete nominal route and provides the reference history, future, and control trace. A decision point is identified immediately before a
safety-relevant expert maneuver. The route is then replayed under the same map, route, scenario, seed, repetition, and synchronous simulation configuration. The online expert controls the ego vehicle before the branch; one candidate
policy is applied for 50 simulator ticks, corresponding to $2.5$~s; and the expert subsequently resumes control. The resulting outcome therefore measures a finite-duration intervention followed by expert recovery. Because the route
is replayed from its beginning rather than restored from a serialized simulator snapshot, we deliberately use the term \emph{matched intervention rollout}. Exact equality of the complete latent simulator state, and hence exact causal identification, is not assumed.

The dataset records synchronized six-view RGB observations, structured ego and route state, navigation information, semantic bird's-eye-view representations, nearby-actor evidence, exact expert and applied controls, and action-specific
future outcomes. Controls are retained at 20~Hz, while visual, structured, and semantic observations are retained at 4~Hz. Semantic BEV and simulator-derived actor evidence are treated as privileged information and are separated from
sensor-oriented model inputs. For learning, all thirteen candidate queries associated with a branch group use the same canonical nominal history. The candidate action is supplied separately through an encoding of its family,
signed magnitude, duration, lateral-control semantics, longitudinal-control semantics, and recovery policy. This construction prevents replay-specific pre-branch differences from serving as action-identifying shortcuts.

A branch-structured dataset is useful only when the branches are collected and matched with sufficient rigor. We therefore audit the source collection at three levels. First, tick-level intervention checks verify the branch start, the complete 50-tick active window, agreement between intended and applied
controls, absence of pre-branch intervention, and return to expert control. Second, replay-fidelity checks compare nominal and intervention histories using ego pose, heading, speed, controls, timing, and available actor-state evidence.
Third, action-leakage classifiers test whether future action identity can be recovered from the pre-branch replay history after metadata and identifiers are removed. Starting from 24,116 raw rollout attempts, the integrity and replay
audits yield a core benchmark containing 606 independent branch groups and 7,878 associated trajectories. All thirteen trajectories from one branch group remain together in the train, validation, or test partition; the branch group,
rather than an individual candidate trajectory or video frame, is the statistical unit.

The benchmark evaluates two related capabilities. The first is
action-conditioned outcome and trajectory prediction:
\begin{equation}
    \left(
        \hat{\mathbf y}_{i,k},
        \hat{\boldsymbol\tau}_{i,k}
    \right)
    =
    f_{\theta}(H_i,e_k).
\end{equation}
The model predicts six continuous outcomes over the intervention horizon: minimum projected time to collision, minimum relevant-actor separation, signed forward displacement, mean horizontal jerk, maximum horizontal acceleration, and maximum absolute yaw rate. It additionally predicts a ten-step future ego trajectory represented by relative position, heading, and speed. Undefined quantities, such as TTC when no relevant closing actor exists, remain masked rather than being converted into artificially safe values. The second capability is complete-bank candidate comparison. Because the executed outcome of every candidate is available within each retained context, the value and regret of a fixed-bank selector can be measured directly, without behavior propensities, importance weighting, or a separately estimated action-value function.

We benchmark a controlled model ladder ranging from a global-mean predictor and restricted action-only and history-only baselines to structured, six-view-visual, multimodal, and privileged-BEV action-conditioned models. On the held-out test partition, the structured history-and-action model
significantly outperforms both restricted baselines for aggregate continuous prediction and future-trajectory error. It reduces macro normalized mean absolute error from $0.6085$ for the action-only model and $0.5826$ for the history-only model to $0.5036$, and reduces average displacement error from
$4.8983$ and $4.2383$~m to $2.2042$~m, respectively. A multimodal RGB, structured-history, and action model obtains the lowest numerical scalar error ($0.5008$), although its advantage over the structured model is not statistically significant and its trajectory error is higher. Action- and
history-shuffling diagnostics further show that the strongest models depend on both candidate semantics and scene context.

The same structured model also supports useful full-information offline selection. Its outcome-derived selector increases balanced continuous policy value from $0.5364$ for the nominal training action prior to $0.5704$ and reduces normalized regret from $0.2674$ to $0.1495$; both paired differences remain significant after Holm correction. This result demonstrates that the prediction gains are sufficiently action-discriminative to improve strict numerical ranking within the fixed bank. It does not establish safe-action selection. In particular, the unconstrained selector narrowly fails a
predeclared minimum-separation non-worsening gate, and a subsequently evaluated global separation-residual guard achieves high empirical coverage only by rejecting every intervention. This negative result exposes a distinction that
is often obscured in decision benchmarks: accurate action-conditioned prediction can support offline ranking without providing sufficiently sharp uncertainty estimates for conservative execution.

The main contributions of this work are:

\begin{itemize}
    \item We introduce a matched multi-action rollout representation in which one canonical decision history is associated with one nominal expert future and twelve physically executed intervention futures with explicit
    control semantics and expert recovery.

    \item We provide a reproducible dataset-validity pipeline covering canonicalization, intervention compliance, modality and temporal completeness, pre-branch replay fidelity, group-disjoint partitioning, and diagnostic testing for future-action leakage.

    \item We establish a benchmark for continuous action-conditioned outcome and future-trajectory prediction across action-only, history-only, structured, visual, multimodal, and privileged representations, showing that joint history--action conditioning is necessary for the strongest
    performance.

    \item We exploit the complete action bank for direct full-information offline policy evaluation, demonstrating significant ranking improvements over a frozen action prior while also showing that a global conservative separation guard can be empirically well covered yet operationally
    vacuous.
\end{itemize}

BranchDrive is therefore positioned as a same-context finite-action decision benchmark rather than a complete offline reinforcement-learning environment or a deployment-ready safety system. Its validated scope is action-conditioned short-horizon prediction and offline comparison among thirteen predefined
policies. Exact causal effects, validated binary unsafe-action selection, replay-noise-equivalent outcomes, and realized closed-loop improvement remain outside the claims of the present work.
\section{Related Work}
\label{sec:related_work}

\subsection{Autonomous-Driving Datasets and Planning Benchmarks}

Large-scale autonomous-driving datasets have established standardized evaluation protocols for perception, tracking, motion forecasting, and planning. nuScenes, the Waymo Open Dataset, and Argoverse~2 provide synchronized multimodal observations, high-definition maps, and dense annotations across diverse real-world environments \cite{caesar2020nuscenes,sun2020waymo,wilson2021argoverse2}. The Waymo Open Motion Dataset further emphasizes interactive forecasting by mining long-duration scenes involving multiple traffic participants \cite{ettinger2021womd}. These resources provide broad coverage of road geometry, appearance, and traffic behavior, but their supervision remains primarily observational: for a given ego-driving context, the dataset records the future realized under the logging policy. Multimodal forecasting targets represent plausible hypotheses about an uncertain future rather than experimentally observed consequences of alternative ego actions.

Planning benchmarks have extended evaluation beyond open-loop displacement error. In particular, nuPlan combines real-world driving logs with a reactive closed-loop simulator and planning-specific measures of progress, comfort, collision avoidance, and rule compliance \cite{caesar2021nuplan}. Different
planners can therefore be executed from a common scenario during evaluation. However, the underlying dataset is not organized as a persistent collection of decision-centered branch groups containing a canonical history, a fixed bank of ego interventions, and the directly observed outcome of every
intervention. BranchDrive addresses a different supervision problem: learning the mapping from one pre-decision context and a specified candidate policy to the resulting short-horizon trajectory and continuous outcome.

\subsection{Safety-Critical Closed-Loop Simulation}

Simulation provides a controlled mechanism for evaluating rare and hazardous driving situations without exposing road users to risk. CARLA supports configurable maps, traffic actors, environmental conditions, sensor suites, and closed-loop vehicle control \cite{dosovitskiy2017carla}. SafeBench builds
on this capability to provide a unified platform for safety-critical scenario generation and autonomous-driving evaluation \cite{xu2022safebench}. Bench2Drive evaluates multiple end-to-end driving capabilities across a broad collection of interactive CARLA scenarios, routes, towns, and environmental
conditions \cite{jia2024bench2drive}. ScenarioNet instead converts heterogeneous real-world driving logs into a unified representation that can be replayed and interacted with in simulation \cite{li2023scenarionet}. These platforms make closed-loop policy evaluation reproducible, but their principal
experimental unit is generally a scenario, route, or complete policy rollout. 

Related methods use perturbations to expose driving policies to behavior not well represented in expert demonstrations. ChauffeurNet, for example, synthesizes perturbed trajectories that create off-route or collision-prone situations and combines them with auxiliary losses encouraging progress and
penalizing undesirable behavior \cite{bansal2019chauffeurnet}. Such augmentation improves robustness to off-distribution states, but it does not provide a reusable outcome table for comparing a fixed set of physically executed ego policies from one decision context. BranchDrive instead varies the ego intervention while holding the intended route, scenario, seed, and branch location fixed. It records exact applied controls and organizes the resulting trajectories into a common branch group, enabling within-context prediction and ranking rather than only route-level policy testing.

\subsection{Reasoning and Alternative-Action Supervision}

A complementary line of research enriches autonomous-driving data with semantic explanations and language-based decision supervision. DriveLM formulates graph visual question answering that connects perception, interaction prediction, and planning questions \cite{sima2024drivelm}. Reason2Drive represents driving reasoning as a sequence of perception, prediction, and decision steps \cite{nie2024reason2drive}, while DriveCoT records CARLA observations, controls, and chain-of-thought labels describing intermediate reasoning and final decisions \cite{wang2024drivecot}. OmniDrive integrates three-dimensional
scene understanding, reasoning, counterfactual questions, and planning within a multimodal driving-agent framework \cite{wang2025omnidrive}.

These datasets provide important supervision for interpretability, question answering, and high-level action reasoning. Their alternative actions are nevertheless expressed primarily through language, decision labels, questions, or synthetically generated annotations. They do not generally
provide the realized closed-loop trajectory and continuous outcome of every candidate action under the same pre-decision context. BranchDrive is therefore complementary: its supervision is consequence-based rather than exclusively
explanatory. Language descriptions may be added downstream, but the primary reference signal is the trajectory produced by executing each intervention in the simulator.

\subsection{Action-Conditioned World Models and Counterfactual Rollouts}

Driving world models seek to predict how observations and traffic states evolve under changes to ego motion, actor behavior, or scene configuration. UniSim constructs a neural closed-loop sensor simulator from recorded driving logs,
supporting novel viewpoints and edited actor configurations
\cite{yang2023unisim}. DriveDreamer learns controllable video generation and future-state prediction from real-world driving data \cite{wang2024drivedreamer}. More recently, ReSim combines real-world expert demonstrations with non-expert simulator behavior to improve generation under hazardous or unusual ego actions, and uses a Video2Reward module to score the generated futures \cite{yang2025resim}. Model-Based Policy Adaptation similarly generates counterfactual trajectories, trains a diffusion-based policy adapter, and employs a learned multi-step value model to select among proposed trajectories \cite{lin2025mpa}.

These approaches demonstrate the utility of action-controllable future generation for planning and policy evaluation. Their alternative futures, however, are produced by a learned simulator or are embedded within a method-specific adaptation pipeline. Consequently, downstream decision error may combine imperfect future synthesis, imperfect reward estimation, and
incorrect action comparison. BranchDrive separates these sources during benchmarking: reference futures are obtained through direct closed-loop CARLA execution and are stored independently of the model being evaluated. A learned predictor may therefore be compared against the same executed future bank
without treating another generative model as ground truth.

\textbf{Positioning of BranchDrive.}
BranchDrive lies at the intersection of driving datasets, closed-loop simulation, intervention-based evaluation, and action-conditioned prediction. Relative to observational datasets, it supplies multiple realized ego futures for each retained decision context. Relative to safety-testing  benchmarks, it organizes trajectories into local branch groups rather than treating complete routes as independent samples. Relative to reasoning datasets, it provides executed outcomes instead of only action labels or linguistic alternatives.
Relative to learned world models, its reference futures are generated by the underlying simulator rather than by the evaluated prediction model. The distinguishing combination is a shared canonical history, a fixed bank of temporally extended intervention policies, direct simulator execution, tick-level treatment records, measured replay fidelity, and branch-group-level evaluation. BranchDrive should consequently be interpreted as a same-context finite-action decision benchmark, not as an unrestricted offline reinforcement-learning dataset or as evidence of exact real-world causal effects.
\section{BranchDrive Dataset}
\label{sec:dataset}

BranchDrive is organized around local driving decisions rather than complete routes or isolated trajectories. For each retained decision context, the dataset associates one canonical pre-decision history with a fixed bank of physically executed ego-vehicle policies. This organization provides direct
supervision for predicting how the same scene evolves under different candidate actions.

\subsection{Branch-Group Representation}
\label{sec:branch_group}

Let $t_i^b$ denote the simulator tick at which the intervention begins for decision context $i$, and let $n_i^b$ be the corresponding saved-observation index. Ego control is represented by
\begin{equation}
    \mathbf{u}_{i,t}
    =
    \begin{bmatrix}
        \delta_{i,t} &
        \alpha_{i,t} &
        b_{i,t}
    \end{bmatrix}^{\top},
\end{equation}
where $\delta$, $\alpha$, and $b$ denote steering, throttle, and braking, respectively.

The canonical model history is obtained from the original nominal expert rollout:
\begin{equation}
\label{eq:canonical_history}
    H_i =
    \left(
        \mathbf{O}^{0}_{i,n_i^b-15:n_i^b},
        \mathbf{U}^{0}_{i,t_i^b-80:t_i^b-1}
    \right).
\end{equation}
Here, $\mathbf{O}^{0}_{i,n_i^b-15:n_i^b}$ contains 16 saved observations at 4~Hz, while $\mathbf{U}^{0}_{i,t_i^b-80:t_i^b-1}$ contains the preceding 80 control commands at 20~Hz. The observation at the branch point is included in $H_i$, but the control applied at $t_i^b$ is excluded. The branch action is
instead supplied separately through its action encoding. This convention prevents the selected intervention from leaking into the common history.

For candidate policy $a_k$, let $e_k$ denote its structured encoding, $\tau_{i,k}$ its simulator-executed future trajectory, $\mathbf{y}_{i,k}$ its continuous outcome vector, and $\mathbf{m}_{i,k}$ its target-availability mask. The atomic dataset unit is the branch group
\begin{equation}
\label{eq:branch_group}
    G_i =
    \left(
        H_i,
        \left\{
            e_k,\tau_{i,k},\mathbf{y}_{i,k},\mathbf{m}_{i,k}
        \right\}_{k=0}^{12},
        q_i
    \right),
\end{equation}
where $q_i$ stores intervention-compliance, replay-fidelity, modality-completeness, and provenance metadata.

Candidate $a_0$ is the nominal expert policy, and $a_1,\ldots,a_{12}$ are alternative intervention policies. Thus, every complete group contains one nominal future and twelve intervention futures. A branch group is treated as
one statistical unit: all thirteen trajectories remain together in the training, validation, or test partition.

\subsection{Two-Pass Collection}
\label{sec:two_pass_collection}

BranchDrive is collected using CARLA in synchronous mode with a simulator interval of $0.05$~s, corresponding to a 20-Hz control rate \cite{dosovitskiy2017carla}. Visual observations, structured measurements, semantic representations, and risk evidence are persisted every five simulator ticks, producing a 4-Hz observation stream.

\begin{figure*}[t]
    \centering
    \includegraphics[width=\textwidth]
    {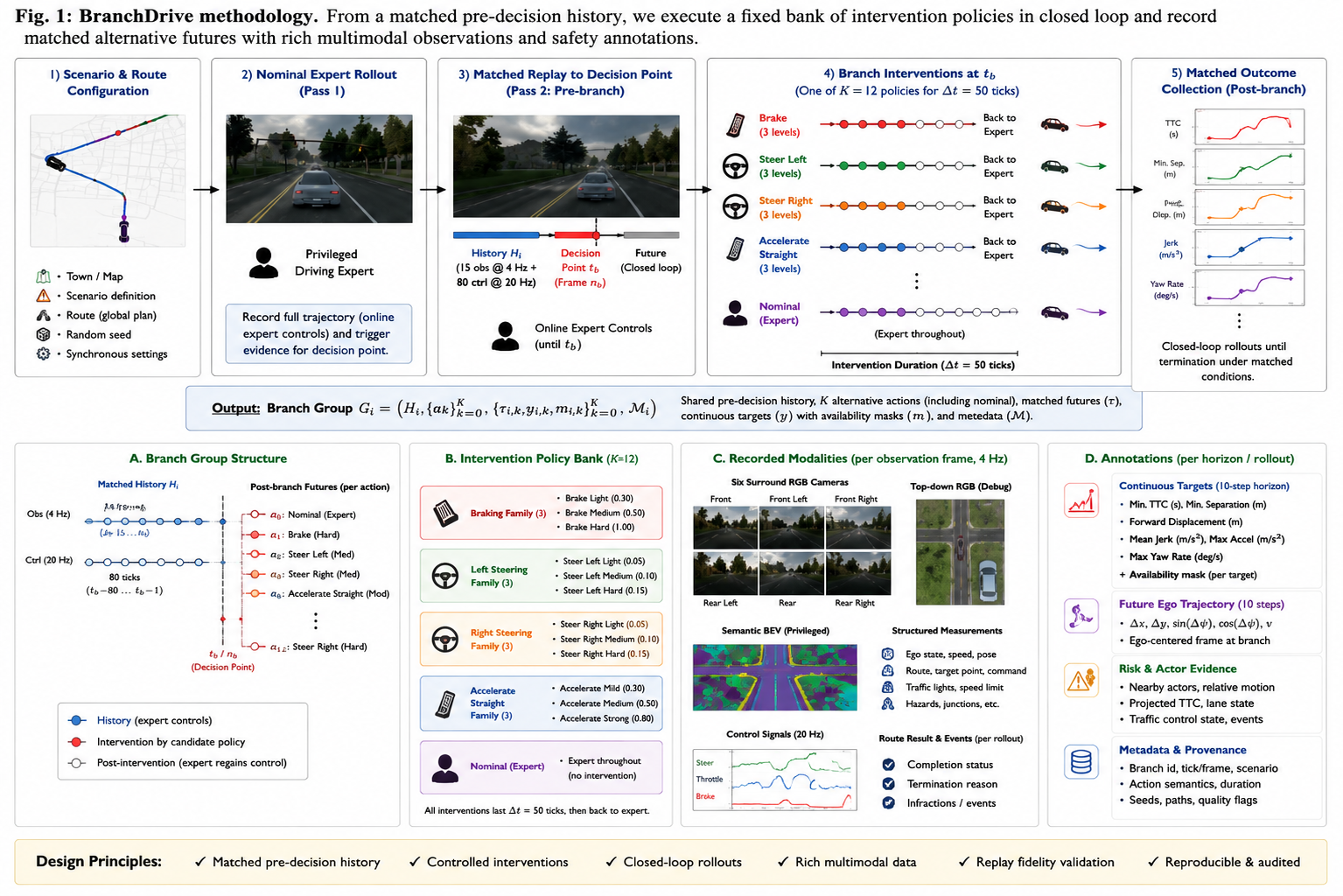}
    \caption{BranchDrive collection and sample construction. A privileged expert first executes the nominal route and supplies the canonical history, original control trace, and branch-trigger evidence. The matched route is then replayed under the same configuration for each of twelve intervention
    policies. The intervention is active for 50 simulator ticks and is followed by online-expert recovery. The resulting branch group contains one shared history, thirteen physically executed futures, action-specific continuous
    targets, and replay-quality metadata.}
    \label{fig:branchdrive_collection}
\end{figure*}

Collection proceeds in two passes, as illustrated in
Fig.~\ref{fig:branchdrive_collection}. In the first pass, a privileged expert executes the complete route. This nominal rollout provides three elements: the canonical history and nominal future, the evidence used to determine the branch point, and the original expert control trace needed by selected
intervention policies. The expert has access to privileged simulator information and is used for data generation and post-intervention recovery; it is not treated as an onboard sensor-only policy.

The branch point is selected from the completed nominal rollout using a scenario-specific expert-guided trigger. The trigger identifies a safety-relevant maneuver and places the branch immediately before the expert response. Trigger selection may inspect the nominal expert future to verify that the maneuver occurred, but no intervention outcome is used to choose the
branch point. The trigger definition, evidence, confidence, branch frame, and simulator tick are retained in the rollout metadata.

In the second pass, the route is replayed once for every intervention policy. For a branch group, the map, route definition, scenario configuration, Traffic Manager seed, repetition index, and synchronous simulator settings
are held fixed. The online expert controls the vehicle before the branch. Candidate policy $a_k$ is then applied for
\begin{equation}
    T_I = 50 \text{ simulator ticks} = 2.5 \text{ s},
\end{equation}
after which control returns to the online expert. The complete branch policy is
\begin{equation}
\label{eq:branch_policy}
    \pi_{i,t}^{(k)}
    =
    \begin{cases}
        \pi_E, & t < t_i^b,\\
        \pi_k, & t_i^b \leq t < t_i^b + T_I,\\
        \pi_E, & t \geq t_i^b + T_I,
    \end{cases}
\end{equation}
where $\pi_E$ is the online expert and $\pi_k$ is the selected intervention policy.

The resulting future therefore measures the consequence of a finite-duration intervention followed by expert recovery. It is not the outcome of applying a constant command indefinitely. In addition, the simulator is replayed from the beginning of the route; the complete latent simulator state is not restored
from a serialized snapshot at $t_i^b$. We consequently use the term \emph{matched intervention rollout} rather than claiming exact counterfactual state identity. Pre-branch replay comparability is evaluated in Sec.~\ref{sec:exp_validity}.

\subsection{Intervention Policy Bank}
\label{sec:action_bank}

The intervention bank contains acceleration, braking, left-steering, and right-steering policies, with three control magnitudes per family. The policies are temporally extended control rules rather than one-time scalar actuator labels. Table~\ref{tab:action_bank} gives their exact semantics.

\begin{table*}[t]
\centering
\caption{Fixed BranchDrive policy bank. Superscript $E$ denotes online-expert control evaluated on the current replay state. Superscript $E,0$ denotes a control copied from the original nominal expert rollout. Every intervention is active for 50 simulator ticks and is followed by online-expert recovery.}
\label{tab:action_bank}
\footnotesize
\setlength{\tabcolsep}{4.2pt}
\begin{tabular}{lllll}
\toprule
Policy family
& Parameter
& Steering command
& Throttle command
& Brake command \\
\midrule
Nominal expert
& ---
& $\delta_t^{E}$
& $\alpha_t^{E}$
& $b_t^{E}$ \\

Accelerate straight
& $\lambda\in\{0.30,0.50,0.80\}$
& $\delta_t=0$
& $\alpha_t=\max(\alpha_t^{E},\lambda)$
& $b_t=0$ \\

Brake
& $\lambda\in\{0.30,0.50,1.00\}$
& $\delta_t=\delta_t^{E}$
& $\alpha_t=0$
& $b_t=\max(b_t^{E},\lambda)$ \\

Steer left
& $\lambda\in\{0.05,0.10,0.15\}$
& $\delta_t=-\lambda$
& $\alpha_t=\alpha_t^{E,0}$
& $b_t=b_t^{E,0}$ \\

Steer right
& $\lambda\in\{0.05,0.10,0.15\}$
& $\delta_t=+\lambda$
& $\alpha_t=\alpha_t^{E,0}$
& $b_t=b_t^{E,0}$ \\
\bottomrule
\end{tabular}
\end{table*}

The accelerate-straight policy suppresses steering and braking while enforcing a minimum throttle level. It therefore changes both lateral and longitudinal behavior and should not be interpreted as an isolated throttle perturbation. The braking policy retains the online expert's steering response on the
replayed state, suppresses throttle, and enforces a minimum braking magnitude.

The steering policies impose an absolute signed steering command while reusing throttle and brake commands stored from the original nominal rollout. These longitudinal controls are nominally matched at the start of the intervention but become open loop with respect to the intervention state after
the trajectories diverge. Because the four families combine forced, original-nominal, and online-expert controls differently, BranchDrive treats them as complete maneuver policies rather than clean one-dimensional actuator
ablations.

The action metadata preserve the canonical action key, policy family, magnitude, duration, steering rule, throttle rule, brake rule, effective action, and recovery controller. This explicit representation allows the control actually applied at every tick to be compared with the intended policy definition.

\subsection{Recorded Modalities}
\label{sec:recorded_modalities}

Each rollout contains synchronized visual observations, structured vehicle and route measurements, privileged spatial representations, exact control records, and frame-level outcome evidence. The principal streams are summarized in Table~\ref{tab:modalities}. 

\begin{table}[t]
\centering
\caption{Principal recorded data streams.}
\label{tab:modalities}
\footnotesize
\setlength{\tabcolsep}{3.0pt}
\begin{tabular}{p{0.34\columnwidth}p{0.12\columnwidth}
                p{0.18\columnwidth}p{0.27\columnwidth}}
\toprule
Stream & Rate & Track & Main contents \\
\midrule
Six surround RGB views
& 4 Hz
& Sensor
& Front, front-left, front-right, rear, rear-left, rear-right \\

Top-down RGB
& 4 Hz
& Debug
& Collection visualization; excluded from sensor-oriented inputs \\

Semantic BEV
& 4 Hz
& Privileged
& Road, lanes, traffic controls, actors, drivable area, and route \\

Ego and route state
& 4 Hz
& Mixed
& Pose, motion, navigation, route-relative and target-point information \\

Actor and risk evidence
& 4 Hz
& Privileged
& Actor distance, relative motion, projected TTC, lane and event evidence \\

Expert and applied controls
& 20 Hz
& Audit
& Online expert, stored nominal expert, and applied steer/throttle/brake \\

Route-level result
& Per rollout
& Evaluation
& Completion status, termination reason, and available infractions \\
\bottomrule
\end{tabular}
\end{table}

\textbf{Surround-view RGB.}
The ego vehicle is equipped with six cameras covering the front, diagonal, and rear sectors. Images are recorded at $1600\times900$ resolution and saved at 4~Hz. A separate top-down camera is retained for visualization and debugging, but is excluded from the sensor-oriented benchmark track.

\textbf{Semantic bird's-eye view.}
The privileged BEV is an ego-centered $256\times256$ raster at 2 pixels/m, covering approximately $128\times128$~m. The raw semantic representation distinguishes road, sidewalk, lane markings, stop signs, traffic-light states, vehicles, and pedestrians. The release also provides binary channels for the
ego vehicle, other vehicles, drivable area, and remaining route. Semantic BEV is reported as privileged information rather than as an onboard sensor input.

\textbf{Structured measurements.}
The measurement stream records ego pose and motion, target speed, speed limit, navigation commands, route target points, route-relative information, hazard indicators, and the applied control. Intervention records additionally retain the online-expert command, the original nominal command when used, branch
timing, action metadata, and the active control source.

\textbf{Risk and actor evidence.}
Frame-level evidence contains ego velocity and acceleration, lane and route state, nearby vehicle and pedestrian statistics, relative motion, projected TTC, traffic-control state, and available event or infraction signals. Undefined evidence remains explicitly unavailable. In particular, the absence
of a relevant closing actor is not converted into an artificially large or automatically safe TTC value.

\textbf{Control audit stream.}
The 20-Hz control log records the intended action, expert command, applied command, control source, and counterfactual-active flag at every simulator tick. It supports independent verification of the intervention start, exact 50-tick duration, control-policy compliance, absence of pre-branch treatment,
and return to the expert controller. 

\subsection{Model Inputs and Targets}
\label{sec:model_inputs_targets}

The current benchmark constructs one model-ready sample per branch group. All thirteen action queries share the canonical nominal history in Eq.~\eqref{eq:canonical_history}. Replay-specific intervention histories are retained for replay-fidelity auditing but are not used as normal prediction
inputs.

The structured history is represented as
\begin{equation}
    \mathbf{X}_i \in \mathbb{R}^{16\times21},
    \qquad
    \mathbf{U}_i \in \mathbb{R}^{80\times3},
\end{equation}
where $\mathbf{X}_i$ contains the 4-Hz branch-centered structured state and $\mathbf{U}_i$ contains the preceding 20-Hz control history. The fixed action bank is represented by
\begin{equation}
    \mathbf{E}\in\mathbb{R}^{13\times29}.
\end{equation}
Each action vector encodes the policy family, signed control magnitude, duration, lateral-control semantics, longitudinal-control semantics, and post-intervention recovery policy. The action encoding is semantic; no free candidate-index identifier is used.

For visual models, the input consists of 16 time steps from the six surround cameras. For the privileged benchmark, the corresponding 16-frame semantic-BEV history is provided separately. Future RGB, future BEV, branch-specific replay histories, future controls, rollout paths, branch  identifiers, and target variables are excluded from model inputs.

The validated scalar target vector is
\begin{equation}
\label{eq:scalar_targets}
    \mathbf{y}_{i,k}
    =
    \begin{bmatrix}
        T^{\min}_{i,k} &
        D^{\min}_{i,k} &
        P^{\parallel}_{i,k} &
        \bar{J}_{i,k} &
        A^{\max}_{i,k} &
        \Omega^{\max}_{i,k}
    \end{bmatrix}^{\top},
\end{equation}
where $T^{\min}$ is minimum projected TTC, $D^{\min}$ is minimum relevant-actor separation, $P^{\parallel}$ is signed forward displacement, $\bar{J}$ is mean horizontal jerk, $A^{\max}$ is maximum horizontal acceleration, and $\Omega^{\max}$ is maximum absolute yaw rate.

Every scalar target is accompanied by a binary availability mask
\begin{equation}
    \mathbf{m}_{i,k}\in\{0,1\}^{6}.
\end{equation}
Unavailable values are excluded from the corresponding loss and metric. This is particularly important for TTC, which is undefined when no relevant closing interaction is observed.

The future ego trajectory contains ten observations following the branch, covering the 2.5-s intervention interval:
\begin{equation}
\label{eq:trajectory_target}
    \tau_{i,k}
    =
    \left\{
    \begin{bmatrix}
        \Delta x_{i,k,j} &
        \Delta y_{i,k,j} &
        \sin\Delta\psi_{i,k,j} &
        \cos\Delta\psi_{i,k,j} &
        v_{i,k,j}
    \end{bmatrix}
    \right\}_{j=1}^{10}.
\end{equation}
Position and heading are expressed in an ego-centered coordinate frame whose origin and orientation are fixed at the branch observation. The first target is the first saved observation after the branch and can therefore contain a
physical response to the control applied at $t_i^b$.

The resulting model-ready arrays have the group-level structure
\begin{equation}
\begin{split}
    &\text{state: } [N,16,21],\qquad
    \text{controls: } [N,80,3],\\
    &\text{actions: } [13,29],\qquad
    \text{scalars: } [N,13,6],\\
    &\text{trajectories: } [N,13,10,5].
\end{split}
\end{equation}
After the collection-integrity and replay-fidelity filtering described in Sec.~\ref{sec:exp_validity}, the core benchmark contains $N=606$ independent branch groups and $7{,}878$ associated trajectories.

The raw release retains additional route-level outcomes and event records. However, the benchmark reported in this paper restricts its headline supervision to the six validated continuous targets and the future ego trajectory. Binary collision, lane-departure, and unsafe-action labels are not
used to support the principal claims.

\subsection{Release Structure}
\label{sec:release_structure}

The release is organized in layers so that the filtering process can be reproduced without modifying the source collection.

\textbf{Raw rollout layer.}
This layer preserves every discovered physical attempt, including completed rollouts, retries, partial attempts, protocol variants, and quarantined records. Each rollout directory contains the available images, measurements,
semantic representations, control log, risk-evidence stream, route result, and collection metadata.

\textbf{Canonical manifest layer.}
A canonical rollout manifest provides one integrity-selected record for each branch-group/action pair and links every intervention to its nominal expert rollout. The manifest records route, scenario, town, seed, repetition, branch
frame and tick, action identity, control semantics, modality paths, and provenance. Nominal linkage records are represented explicitly rather than being inferred from directory names.

\textbf{Quality-control layer.}
Rollout- and group-level tables store intervention-compliance results, temporal-window completeness, modality availability, replay discrepancies, structural eligibility, exclusion reasons, and action-leakage diagnostics. The raw data are therefore not silently relabeled as benchmark-valid; the
release distinguishes source attempts from the filtered benchmark population.

\textbf{Model-ready layer.}
The model-data package contains the frozen group split, canonical histories, candidate action matrix, scalar targets, target masks, future trajectories, RGB and BEV path indices, candidate ordering, feature schemas, and training-only normalization statistics. All joins are keyed by branch-group
identity and canonical action key rather than by positional directory order.

\textbf{Benchmark and reproducibility layer.}
The release additionally provides target-generation code, baseline implementations, evaluation scripts, frozen configurations, trained checkpoints where permitted, prediction files, and SHA-256 checksums. Separate sensor-oriented and privileged input contracts prevent semantic BEV, simulator actor state, or retrospective trigger information from being
inadvertently presented as onboard observations.

The final train/validation/test partition contains 431, 88, and 87 branch groups, respectively. Since each group contains thirteen dependent futures, the effective sample count is the number of branch groups rather than the number of trajectories or video frames.
\section{Benchmark Protocol}
\label{sec:benchmark_protocol}

\subsection{Evaluation population.}
All experiments use the frozen core replay-valid population of 606 branch groups and 7,878 associated trajectories. Each group contains one nominal expert future and twelve intervention futures. Groups are partitioned into 431/88/87 train/validation/test contexts, corresponding to
5,603/1,144/1,131 candidate trajectories. All thirteen candidates from one context remain in the same partition, and the branch group is the independent unit for training splits, uncertainty estimates, and hypothesis tests.

\subsection{Model input.}
Every candidate query uses one shared canonical nominal history comprising 16 structured observations at 4 Hz and 80 past control commands at 20 Hz. The applied branch action is excluded from this history. Candidate policies
are supplied separately through a 29-dimensional encoding of maneuver family, signed magnitude, duration, lateral-control semantics, longitudinal-control semantics, and expert recovery. RGB models use sixteen time steps from six surround cameras; the privileged model uses the corresponding semantic-BEV
history. Replay-specific intervention histories, file paths, branch identifiers, future controls, and post-branch observations are never used as model inputs.

\subsection{Prediction tasks.}
For each history--action pair, the model predicts six 2.5-s scalar outcomes: minimum TTC, minimum actor separation, signed forward displacement, mean horizontal jerk, maximum horizontal acceleration, and maximum absolute yaw rate. It also predicts a ten-step future ego trajectory represented by
$(x,y,\sin\Delta\psi,\cos\Delta\psi,v)$. Undefined TTC values are retained as unavailable and excluded through a target mask rather than being interpreted as safe observations.

\subsection{Compared models.}
The benchmark includes a training-mean baseline (M0), an action-only model (M1), a history-only model (M2), structured history plus action (M3), six-view RGB plus action (M4), RGB plus structured history plus action (M5), and semantic BEV plus action (M6). M6 is a privileged comparator and is
reported separately from sensor-oriented models.

\subsection{Training and test locking.}
Continuous targets are normalized using training-set statistics only. M1--M6 are trained using three seeds, with checkpoint selection and early stopping performed exclusively on validation data. The model architecture, checkpoint identities, candidate ordering, normalization, metric definitions,
and ranking utility are locked before test outcomes are accessed. Test evaluation is performed once under the frozen configuration.

\subsection{Metrics and statistics.}
Scalar outcomes are evaluated using MAE, RMSE, NMAE, and Spearman correlation; trajectories are evaluated using ADE, FDE, heading MAE, and speed MAE. The principal scalar score is the equal-weight mean of the six target NMAEs. Confidence intervals use 2,000 bootstrap resamples of complete
branch groups. Paired comparisons use 10,000 branch-level permutations with Holm--Bonferroni correction. Seed predictions are averaged within each branch before statistical comparison.

\subsection{Offline candidate ranking.}
Predicted outcomes are converted to a utility whose normalization is fitted using training targets only. Under the principal balanced profile, minimum separation, signed displacement, and the aggregate comfort term receive equal
category weight. Because every one of the thirteen candidate outcomes is observed in each retained context, the value of a selected action is measured by direct lookup of its executed outcome. The fixed-bank oracle is therefore
the best of the thirteen tested policies, not a global driving oracle.

\subsection{Scope.}
The validated benchmark addresses continuous outcome prediction, future ego trajectory prediction, and strict numerical fixed-bank ranking. Binary unsafe-action selection, replay-noise-equivalent action classes, causal effect
identification, and realized closed-loop improvement are not evaluated.

\begin{table}[t]
\centering
\caption{Frozen BranchDrive benchmark protocol.}
\label{tab:benchmark_protocol}
\footnotesize
\setlength{\tabcolsep}{3.2pt}
\begin{tabular}{p{0.29\columnwidth}p{0.63\columnwidth}}
\toprule
Item & Frozen protocol \\
\midrule
Statistical unit
& One complete branch group containing 13 candidate futures \\

Population
& 606 groups; 431/88/87 train/validation/test \\

Canonical history
& 16 state frames at 4 Hz and 80 control ticks at 20 Hz \\

Candidate bank
& Nominal expert plus 12 interventions in four families and three
magnitudes per family \\

Action encoding
& 29-D family, signed magnitude, duration, control semantics, and recovery \\

Prediction horizon
& Ten saved frames over the 2.5-s intervention interval \\

Scalar targets
& TTC, separation, displacement, jerk, acceleration, yaw rate \\

Trajectory target
& $(x,y,\sin\Delta\psi,\cos\Delta\psi,v)$ at ten future steps \\

Normalization
& Training partition only; unavailable TTC values masked \\

Model selection
& Validation only; three seeds for M1--M6 \\

Test protocol
& Model and metric lock followed by one-time test evaluation \\

Statistics
& Group bootstrap and paired group-level permutation tests \\

Ranking
& Frozen continuous utility with direct observed-outcome lookup \\

Excluded claims
& Binary safety, null-equivalence, causality, and closed-loop benefit \\
\bottomrule
\end{tabular}
\end{table}
\section{Experiments}
\label{sec:experiments}

\subsection{Research Questions and Experimental Protocol}
\label{sec:exp_protocol}

We evaluate BranchDrive at four progressively stronger levels. \emph{RQ1} asks whether the collected branches satisfy the intervention protocol, replay-matching criteria, and pre-branch leakage checks. \emph{RQ2} asks whether a model conditioned jointly on the driving history and a candidate intervention predicts short-horizon outcomes and ego motion more accurately than action-only or history-only alternatives. \emph{RQ3} asks whether these predictions support better fixed-bank decisions under full-information offline evaluation. \emph{RQ4} tests whether a validation-calibrated minimum-separation guard remains useful rather than becoming vacuous. The experiments deliberately stop at offline evaluation: binary unsafe-action labels, replay-calibrated outcome equivalence, and realized closed-loop improvement are not claimed.

\textbf{Population and split.}
The source audit covered 24,116 rollout attempts from 38 scenario types in Town12 and Town13. After control-compliance, modality, and replay-fidelity filtering, the principal benchmark contains 606 independent branch groups and 7,878 associated trajectories. The final population spans nine scenario types, two towns, and 167 route-geometry clusters. Each group contains one nominal expert future and twelve intervention futures; all thirteen candidates remain in the same partition. The frozen split contains 431/88/87 train/validation/test groups, corresponding to 5,603/1,144/1,131 candidate trajectories. The branch group, rather than an individual candidate trajectory, is the independent statistical unit.

\textbf{Inputs and targets.}
Every candidate query uses the same canonical nominal history. The structured input comprises 16 saved state observations with 21 variables and 80 control ticks with steering, throttle, and brake. Candidate policies are represented by a 29-dimensional encoding of family, signed magnitude, duration, lateral and longitudinal control semantics, and expert recovery. Visual models receive 16 time steps from six cameras, represented by frozen 512-dimensional ResNet-18 features per view; the privileged model receives four semantic-BEV channels at $256\!\times\!256$ over the same history. Each query predicts six scalar outcomes---minimum TTC, minimum separation, signed forward displacement, mean horizontal jerk, maximum horizontal acceleration, and maximum absolute yaw rate---and a ten-step ego trajectory $(x,y,\sin\Delta\psi,\cos\Delta\psi,v)$. TTC is numeric for 7,763 candidates and unavailable for 115 candidates with no relevant closing actor; unavailable values remain masked.

\textbf{Models and training.}
M0 predicts the global training mean; M1 uses the candidate action only; M2 uses structured history only and broadcasts one prediction across candidates; M3 fuses structured history and action; M4 fuses six-view RGB history and action; M5 combines RGB, structured history, and action; and M6 uses privileged semantic BEV and action. M1--M6 were trained for seeds 11, 23, and 37 using AdamW, cosine decay with five warm-up epochs, at most 150 epochs, and validation early stopping with patience 20. The loss is the sum of target-wise masked Huber loss, trajectory Smooth-$L_1$ loss, and a $0.05$ heading-unit-norm penalty. All normalization statistics are computed from training groups only. M4 and M5 use a frozen ImageNet ResNet-18; no stochastic image augmentation is applied. Training used PyTorch 2.5.1/CUDA 12.1 on NVIDIA A40 nodes. The locked group batch sizes were 64 for M1--M3, 16 for M4--M5, and 2 for M6.

\textbf{Metrics and statistics.}
Scalar prediction is evaluated using MAE, RMSE, normalized MAE (NMAE), and Spearman correlation; trajectory prediction uses ADE, FDE, heading MAE, and speed MAE. The principal scalar score is the equal-weight mean of the six target NMAEs. Learned-model tables report the mean over three seeds. Confidence intervals use 2,000 bootstrap resamples of complete branch groups. Paired model comparisons use 10,000 branch-level permutations with Holm--Bonferroni correction. For paired tests, errors are first averaged across seeds within each branch group. The test targets and ranking outcomes were accessed only after the corresponding model and protocol locks were frozen.

\begin{table*}[t]
\centering
\caption{Dataset qualification and frozen benchmark population. ``Traj.'' denotes associated nominal and intervention trajectories; it is not an independent-sample count. The primary experiments use the core replay-valid population.}
\label{tab:population}
\footnotesize
\setlength{\tabcolsep}{5.2pt}
\begin{tabular}{lrrrrr}
\toprule
Stage or population & Train & Validation & Test & Total groups & Traj. \\
\midrule
Discovered branch-group identifiers & -- & -- & -- & 2,467 & -- \\
Exact thirteen-action banks          & -- & -- & -- & 1,751 & 22,763 \\
All-controls-valid banks             & -- & -- & -- & 649   & 8,437 \\
Integrity-complete banks             & 444 & 95 & 95 & 634 & 8,242 \\
\textbf{Core replay-valid}           & \textbf{431} & \textbf{88} & \textbf{87} & \textbf{606} & \textbf{7,878} \\
Extended replay-valid                & 427 & 88 & 85 & 600 & 7,800 \\
\bottomrule
\end{tabular}
\vspace{2pt}

\parbox{0.97\textwidth}{\scriptsize
The raw collection includes 494 partial/quarantined attempts and 402 duplicate attempts. Canonicalization produced 24,132 branch-associated records: 22,312 physical records and 1,820 explicit nominal links, with no unexplained provenance. All 1,902 acceleration candidates in the 634 integrity-complete banks use the locked zero-steering accelerate-straight protocol.}
\end{table*}

\begin{figure*}[t]
\centering
\subfloat[Qualification attrition.]{\includegraphics[width=0.238\textwidth]{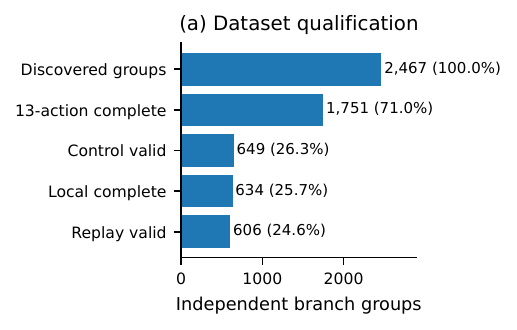}}\hfill
\subfloat[Action-leakage diagnostic.]{\includegraphics[width=0.238\textwidth]{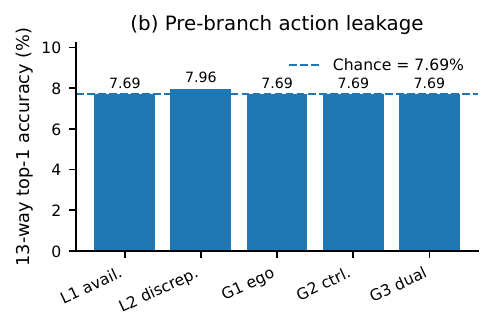}}\hfill
\subfloat[Scalar prediction.]{\includegraphics[width=0.238\textwidth]{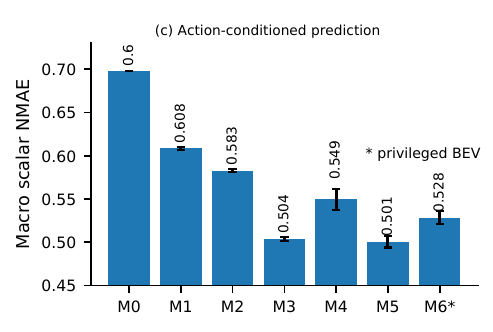}}\hfill
\subfloat[Offline policy value.]{\includegraphics[width=0.238\textwidth]{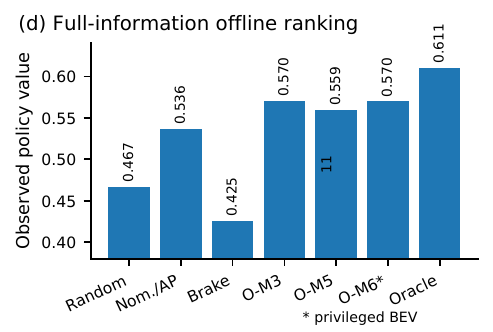}}
\caption{Principal experimental results. (a) The dominant dataset reduction arises from enforcing exact intervention-control semantics. (b) Real-data 13-way leakage classifiers remain at chance; deliberately injected positive controls achieve at least 96.2\% accuracy. (c) History-plus-action models provide the best aggregate scalar prediction. Error bars show standard deviation across three seeds. (d) O-M3 improves full-information balanced-utility value over the nominal/action-prior baseline. M6 and O-M6 use privileged semantic BEV.}
\label{fig:main_results}
\end{figure*}

\subsection{Collection Integrity, Replay Fidelity, and Leakage}
\label{sec:exp_validity}

\textbf{Collection integrity.}
All attempts were audited without modifying the source collection. The checks covered monotonic and unique simulator ticks, the recorded branch start, absence of pre-branch intervention, exactly 50 active intervention ticks, action-specific steering/throttle/brake semantics, expert recovery at the following tick, the complete structured and control histories, the 2.5-s future, risk evidence, and all six RGB streams. Of 2,467 discovered branch groups, 1,751 contained the exact thirteen-action bank, 649 passed all control checks, and 634 additionally contained the required local observations. The major attrition therefore results from treatment noncompliance rather than missing local data. In particular, some newer acceleration attempts retained online-expert steering instead of the locked zero-steering policy; they remain in the raw release but are excluded from the strict bank.

\textbf{Replay fidelity.}
Replay tolerances were fixed from 442 structurally eligible training groups; two additional training groups were excluded because their nominal and intervention metadata contained confirmed seed disagreement. For each replay variable, the threshold is the empirical 99th percentile (higher order statistic) of the maximum discrepancy across the twelve intervention pairs in a training group. No nearest-pedestrian-distance threshold was fabricated because training support was absent. On validation, 1,116/1,140 pairs (97.89\%) and 88/95 complete banks (92.63\%) passed the core criteria. The sealed test evaluation retained 1,078/1,140 core pairs (94.56\%, 95\% CI 90.44--98.07\%) and 87/95 complete banks (91.58\%, 95\% CI 86.29--96.84\%). Extended criteria retained 1,067 pairs (93.60\%) and 85 groups (89.47\%). Thus, the predeclared 98\% pair-retention target was not met, whereas the 90\% core group-retention criterion was met. The resulting modeling population contains the 606 core-valid groups in Table~\ref{tab:population}.

\textbf{Action leakage.}
We next attempted to infer future action identity using only replay-specific pre-branch availability, discrepancy features, ego history, control history, or their combination; action metadata, identifiers, paths, post-branch rows, and outcomes were prohibited. Chance is $1/13=7.692\%$. Test top-1 accuracy was 7.692\% for the availability classifier and all three GRUs, and 7.958\% for the discrepancy logistic regression. The latter had a branch-bootstrap interval of 7.692--8.311\% and was not significant after correction ($p_{\mathrm{Holm}}=0.288$). Explicit-label and synthetic-offset positive controls achieved at least 96.2\%, while random-label controls remained at chance. We therefore found no detectable action leakage under the tested representations; this is a diagnostic result, not a proof that no untested leakage channel exists.

\subsection{Action-Conditioned Outcome and Trajectory Prediction}
\label{sec:exp_prediction}

Table~\ref{tab:prediction} reports held-out performance on 87 branch groups. M3 is the strongest parsimonious model: compared with M1, it reduces branch-level macro NMAE by 0.1049 (95\% CI $[-0.1265,-0.0821]$) and ADE by 2.694 m (95\% CI $[-3.192,-2.188]$); compared with M2, the reductions are 0.0791 (95\% CI $[-0.0950,-0.0627]$) and 2.034 m (95\% CI $[-2.263,-1.800]$). All four comparisons remain significant after Holm correction ($p_{\mathrm{Holm}}=9\times10^{-4}$). These results establish that neither an action prior nor scene history alone explains the predictive performance.

M5 achieves the lowest numerical scalar macro NMAE, but its difference from M3 is small and nonsignificant ($-0.0028$, 95\% CI $[-0.0129,0.0069]$, $p_{\mathrm{Holm}}=0.587$). M3 is better on every trajectory metric; M5 increases ADE by 0.330 m relative to M3 (95\% CI $[0.142,0.511]$). RGB alone (M4) does not match structured history (M3), and the privileged BEV model (M6) does not outperform M5. Accordingly, M5 is the best numerical aggregate scalar model, while M3 is the best trajectory predictor and the strongest efficiency--accuracy trade-off.

The target-level results expose an important limitation. M3 is best on minimum separation, displacement, and yaw rate, whereas M5 is best on jerk and maximum acceleration. All learned models obtain TTC NMAE near 1.10; TTC prediction is therefore substantially weaker than the other outcomes and is not used as a hard safety filter. Test-time intervention shuffling increases M3 macro NMAE by 0.146 and ADE by 3.051 m; history shuffling increases them by 0.195 and 4.470 m; shuffling both increases them by 0.249 and 5.712 m. M5 exhibits the same qualitative dependence. The median predicted-to-observed candidate-spread ratio is 0.647 for M3 and 0.558 for M5, showing that both distinguish candidates but compress some within-context action variation. Full 96-image decoding and frozen ResNet inference require approximately 1.37 s per sample, so the current visual pipeline is an offline benchmark rather than a real-time deployment implementation.

\begin{table*}[t]
\centering
\caption{Action-conditioned prediction on the frozen test split. Learned-model values are means over seeds 11, 23, and 37; M0 is deterministic. Lower is better. $^{\dagger}$M6 uses privileged semantic BEV.}
\label{tab:prediction}
\scriptsize
\setlength{\tabcolsep}{3.6pt}
\begin{tabular}{llrrrrrr}
\toprule
Model & Input & Params. (M) & Macro NMAE & ADE (m) & FDE (m) & Heading ($^\circ$) & Speed (m/s) \\
\midrule
M0 & Mean                         & 0     & 0.6976 & 5.8227 & 10.2097 & 15.0468 & 4.9206 \\
M1 & Action                       & 0.202 & 0.6085 & 4.8983 &  7.9022 &  8.9867 & 4.2353 \\
M2 & State/control history        & 0.398 & 0.5826 & 4.2383 &  8.8550 & 14.7979 & 3.7647 \\
M3 & State/control + action       & 0.553 & 0.5036 & \textbf{2.2042} & \textbf{4.3739} & \textbf{5.9238} & \textbf{2.3022} \\
M4 & RGB + action                 & 1.717 & 0.5495 & 3.5473 &  6.1814 &  8.4443 & 3.1534 \\
M5 & RGB + state/control + action & 2.333 & \textbf{0.5008} & 2.5346 & 4.6759 & 6.9289 & 2.4524 \\
M6$^{\dagger}$ & Semantic BEV + action & 1.448 & 0.5283 & 3.1053 & 5.1927 & 7.2182 & 2.7100 \\
\bottomrule
\end{tabular}
\end{table*}

\subsection{Complete-Bank Ranking and Offline Policy Evaluation}
\label{sec:exp_ranking}

We derive candidate rankings from predicted primitive outcomes rather than training a separate ranker. Each primitive is transformed to $[0,1]$ using the training-set 5th and 95th percentiles, with orientation chosen so that larger desirability is better. The predeclared balanced utility is
\begin{equation}
U=\frac{1}{3}d_{\mathrm{sep}}+\frac{1}{3}d_{\mathrm{disp}}+
\frac{1}{9}\left(d_{\mathrm{jerk}}+d_{\mathrm{acc}}+d_{\mathrm{yaw}}\right).
\label{eq:balanced_utility}
\end{equation}
TTC and unvalidated binary event labels are excluded. Because every candidate has a directly observed outcome in each test group, the value of a selected fixed-bank action is measured by direct lookup rather than importance weighting or an estimated $Q$-function.

Table~\ref{tab:ranking} shows the principal balanced-utility results. The training ActionPrior selects nominal expert for all profiles and is therefore identical to the nominal baseline; the validation-selected fixed brake is brake $+0.30$. O-M3 raises policy value from 0.5364 to 0.5704 and reduces normalized regret from 0.2674 to 0.1495. The paired differences relative to ActionPrior are $+0.0341$ for value and $-0.1179$ for regret, with $p_{\mathrm{Holm}}=0.0024$ for both. O-M3 captures approximately 46\% of the numerical gap between nominal and the fixed-bank oracle. It selects 8--10 distinct actions across seeds, with mean pairwise seed agreement 0.736, indicating context-dependent but not perfectly stable decisions. O-M5 reduces regret but its policy-value improvement does not survive correction ($p_{\mathrm{Holm}}=0.102$), and privileged O-M6 does not significantly outperform O-M5.

These results validate offline candidate comparison, not safe-action selection. The utility contains continuous separation, progress, and comfort terms; the fixed-bank oracle is only the best of thirteen tested policies; and no replay-derived outcome-equivalence tolerance is available. We therefore report strict numerical ranking and retain the primitive selected outcomes in the supplementary tables.

\begin{table}[t]
\centering
\caption{Full-information offline policy evaluation on 87 test groups under the balanced utility. Higher is better except NRegret. $^{\dagger}$Privileged BEV.}
\label{tab:ranking}
\setlength{\tabcolsep}{2.1pt}
\resizebox{\columnwidth}{!}{%
\begin{tabular}{lrrrrrr}
\toprule
Selector & Value & $\Delta$Nom. & NReg. & Pair & NDCG & Top-1 \\
\midrule
Random                & 0.4669 & $-0.0694$ & 0.4659 & 0.5000 & 0.9309 & 0.0813 \\
Nominal / ActionPrior & 0.5364 &  0.0000   & 0.2674 & 0.5263 & 0.9475 & 0.1264 \\
AlwaysBrake           & 0.4255 & $-0.1109$ & 0.6148 & 0.4820 & 0.9222 & 0.0115 \\
O-M3                  & \textbf{0.5704} & \textbf{$+0.0341$} & 0.1495 & 0.7340 & \textbf{0.9772} & 0.3946 \\
O-M5                  & 0.5594 & $+0.0230$ & 0.1615 & \textbf{0.7435} & 0.9741 & 0.3602 \\
O-M6$^{\dagger}$      & 0.5698 & $+0.0335$ & \textbf{0.1441} & 0.7432 & 0.9764 & \textbf{0.4674} \\
Fixed-bank oracle     & 0.6105 & $+0.0741$ & 0.0000 & 1.0000 & 1.0000 & 1.0000 \\
\bottomrule
\end{tabular}}
\end{table}

\subsection{Conservative Calibration and Deployment Boundary}
\label{sec:exp_calibration}

The positive offline ranking result did not satisfy the deployment-readiness audit. Relative to ActionPrior, the parent O-M3 selections reduced mean minimum separation by 0.033 m; the frozen non-worsening gate therefore failed even though aggregate utility and regret improved. The practical materiality of 3.3 cm cannot be established because repeated nominal outcome replays were not collected, but changing the gate after observing the result would be post hoc.

We consequently evaluated a validation-calibrated one-sided separation guard. The guard used the 95th-percentile finite-sample order statistic of each validation group's maximum overprediction residual across its twelve interventions. The resulting margin was 7.544 m. It achieved 99.62\% candidatewise and 95.45\% groupwise empirical coverage, but rejected every intervention: the selector chose nominal in all 88 validation groups, had one unique action, and obtained the same policy value as ActionPrior (0.55226). The guard is therefore conservative but operationally vacuous. No CARLA execution was launched, and no safety or formal coverage claim is made.

\subsection{Experimental Conclusions}
\label{sec:exp_conclusions}

The experiments support three bounded conclusions. First, the audited release provides 606 replay-valid, leakage-screened decision contexts with complete thirteen-action supervision. Second, action-conditioned models that combine history and intervention encoding substantially outperform action-only and history-only predictors; M3 is the strongest trajectory model, while M5 has a statistically indistinguishable numerical advantage on aggregate scalar NMAE. Third, O-M3 converts these predictions into a significant full-information offline ranking improvement over a frozen action prior. The failed separation guard establishes the boundary of the present evidence: calibrated conservative execution remains unresolved, and the paper makes no claim of closed-loop or real-world safety improvement.

\section{Discussion, Limitations, and Intended Use}
\label{sec:discussion}

\subsection{Main Findings}
\label{sec:discussion_findings}

The principal contribution of BranchDrive is the organization of driving data around matched decision contexts rather than isolated trajectories. After collection-integrity and replay-fidelity filtering, the benchmark contains 606 independent branch groups and 7,878 associated trajectories. Each group
provides one canonical pre-decision history, one nominal expert future, and twelve simulator-executed intervention futures. This structure makes it possible to supervise and evaluate the mapping
\begin{equation}
    (H_i,e_k)
    \longrightarrow
    \left(
        \mathbf{y}_{i,k},
        \tau_{i,k}
    \right)
\end{equation}
for several candidate policies under the same intended decision context. The effective sample size is therefore 606 branch groups rather than 7,878 independent trajectories.

The dataset-validity experiments show that this structure cannot be assumed from directory counts alone. Only 634 of the 2,467 discovered branch groups satisfied the complete collection-integrity requirements, and replay-fidelity filtering reduced the final core population to 606 groups. On the sealed test split, 1,078 of 1,140 nominal--intervention pairs satisfied the frozen core replay criteria, corresponding to a pair-retention rate of 94.56\%. This rate was below the predeclared 98\% operational target, whereas 87 of 95 complete
action banks passed, yielding a group-retention rate of 91.58\% and exceeding the predeclared 90\% complete-bank criterion. The benchmark therefore retains only complete action banks that satisfy the frozen group-level rule, while the less favorable pair-level result remains explicitly reported.

The action-leakage experiment provides additional evidence that the retained branch-specific histories do not trivially reveal which policy will be executed. Logistic classifiers using replay availability or discrepancy features and recurrent classifiers using ego and control histories performed at or near chance on both the 13-way and intervention-only 12-way tasks. The smallest Holm-adjusted permutation-test value was 0.288, while explicit-label and synthetic-offset positive controls reached approximately 96--100\% accuracy. The supported conclusion is therefore that no future-action identity
was detected by the evaluated diagnostics, rather than the stronger claim that all possible leakage channels have been eliminated.

The prediction experiments demonstrate that both scene context and candidate policy information are required. The action-only model M1 achieved a scalar macro NMAE of 0.6085 and trajectory ADE of 4.8983~m, while the history-only model M2 obtained 0.5826 and 4.2383~m. Combining structured history and action
in M3 reduced these errors to 0.5036 and 2.2042~m, respectively. All four principal M3 comparisons against M1 and M2 remained significant after Holm correction with adjusted $p=0.0009$. Action and history shuffling also degraded M3, supporting the interpretation that its performance depends on
the interaction between the current context and the proposed intervention rather than on a global action prior or context difficulty alone.

The multimodal M5 model obtained the lowest numerical scalar macro NMAE, 0.5008, but its difference from M3 was not statistically significant ($p=0.5867$), and its trajectory errors were higher. M3 therefore provides the strongest parsimonious result and the best trajectory prediction, while
M5 provides the lowest numerical aggregate scalar error. Minimum-TTC prediction remained difficult for every model, with target-level NMAE near 1.10. TTC should consequently not be treated as a reliable standalone filter under the current formulation.

The complete branch bank also enables direct offline evaluation of outcome-derived selectors. Under the frozen balanced continuous utility, O-M3 increased policy value from 0.5364 for the nominal ActionPrior to 0.5704 and reduced normalized regret from 0.2674 to 0.1495. Both differences remained significant after Holm correction with adjusted $p=0.0024$. O-M3 also improved pairwise concordance from 0.5263 to 0.7340 and strict top-1 selection from 0.1264 to 0.3946. Since the outcome of every fixed-bank candidate is directly available, these results require neither behavior-policy propensity estimates nor a separately learned action-value function.

The deployment-oriented analyses nevertheless show that improved aggregate prediction and ranking do not establish conservative action selection. The unconstrained parent O-M3 selector produced a mean selected minimum separation 0.033~m below ActionPrior and therefore failed the predeclared non-worsening gate. A subsequent validation-calibrated one-sided separation guard produced a global residual margin of 7.544~m. Although it achieved 99.62\% candidatewise and 95.45\% groupwise empirical coverage, it rejected every intervention and
collapsed to nominal selection. Its policy value consequently equaled the ActionPrior value. This result is scientifically informative: a bound can exhibit high empirical coverage while being too broad to support any useful decision.

Taken together, the experiments establish three progressively stronger but distinct conclusions:
\begin{enumerate}
    \item branch-structured supervision supports action-conditioned short-horizon prediction;
    \item the resulting predictions support useful strict numerical ranking within the fixed action bank;
    \item the evaluated global uncertainty guard is not sufficiently sharp for conservative intervention selection.
\end{enumerate}
The third result prevents the first two from being interpreted as evidence of deployment readiness.

\subsection{Modality Interpretation}
\label{sec:discussion_modalities}

Structured ego, route, and control histories were highly informative for the evaluated targets. M3 substantially outperformed both restricted baselines and also produced the lowest ADE, FDE, heading error, and speed error among the
evaluated models. This indicates that recent vehicle dynamics, route-relative state, navigation information, and past controls provide a strong representation for predicting the consequences of the short, explicitly defined candidate policies.

RGB input provided more limited incremental benefit under the evaluated implementation. M4, which used frozen six-view RGB features and candidate action without structured history, was weaker than M3 on both scalar and trajectory metrics. M5 combined RGB, structured history, and action and slightly improved aggregate scalar NMAE, but not significantly, while
increasing trajectory error. This result does not establish that visual information is unnecessary. It establishes that the particular frozen ResNet-18 feature representation and temporal fusion used here did not provide a statistically supported advantage over structured history for the principal scalar score. More object-centric visual representations, trainable visual backbones, or explicit actor-motion supervision may change this comparison. 

The semantic-BEV model M6 should be interpreted as a privileged comparator rather than a guaranteed upper bound. It obtained scalar macro NMAE of 0.5283 and ADE of 3.1053~m and did not outperform M3 or M5. Direct access to semantic road and occupancy structure does not by itself guarantee superior prediction when the model lacks some explicit ego-dynamics information or when the representation and temporal encoder are not optimally matched to the target. The M6 result therefore measures one privileged architecture, not the maximum
achievable performance with privileged simulator state.

Prediction-spread diagnostics reveal a further distinction between average regression accuracy and candidate discrimination. The median ratio of predicted to observed within-group candidate variance was 0.647 for M3 and
0.558 for M5. Both models therefore compressed the differences among candidate outcomes, although M3 was less underdispersed. Such compression can preserve reasonable average error while reducing the margin between competing actions, which helps explain why ranking performance does not follow scalar prediction error exactly.

Computational cost also differs substantially across modalities. The complete M4/M5 visual path required  approximately 1.37~s per branch query when decoding 96 images and computing frozen ResNet features, whereas the structured three-seed O-M3 selector required approximately 13~ms on CPU. The current visual implementation is therefore suitable for offline benchmarking but not for a high-frequency closed-loop decision module without incremental feature caching, a lighter visual encoder, or asynchronous inference. 

\subsection{Threats to Validity}
\label{sec:discussion_limitations}

\textbf{Matched replay is not exact state restoration.}
Intervention branches are reconstructed by rerunning the route under the same route, scenario, seed, repetition, and synchronous simulation settings. The full CARLA, Traffic Manager, actor-controller, sensor-queue, and random
generator state is not restored from a serialized branch snapshot. Residual differences may therefore remain in surrounding-actor trajectories, traffic-control phases, controller state, and sensor timing. The replay audit measures and filters these discrepancies but cannot make the alternatives
exact latent-state counterfactuals. The results should consequently be interpreted as comparisons among quality-controlled matched intervention rollouts rather than as exact causal treatment effects. 

\textbf{Retrospective decision-point selection.}
The branch detector uses the completed nominal expert future to verify that a safety-relevant maneuver occurred. The retained population is therefore conditioned on contexts in which the expert exhibited a recognizable response. Contexts involving unrecognized strategies, aborted maneuvers, or nominal expert failures may be underrepresented. BranchDrive evaluates action
consequences conditional on the trigger and admission protocol; it does not represent the complete distribution of hazardous driving encounters. 

\textbf{Intervention-policy asymmetry.}
The four action families do not isolate individual actuator dimensions. Accelerate-straight sets steering to zero, suppresses braking, and enforces a minimum throttle. Braking preserves online-expert steering while suppressing throttle and enforcing a minimum brake command. Steering interventions impose
absolute lateral commands while using longitudinal controls from the original nominal expert trace. Differences among families therefore reflect complete maneuver policies and their feedback structures, not clean one-variable actuator perturbations.

\textbf{Short horizon and expert recovery.}
The reported prediction targets cover the 2.5-s forced-intervention interval. The collection policy returns control to the online expert after 50 simulator ticks. Longer-horizon consequences would therefore depend on the privileged recovery policy, while actions whose advantages require longer execution may be disadvantaged by the fixed intervention duration. The present claims are restricted to short-horizon prediction and comparison under this fixed policy contract.

\textbf{Finite candidate support.}
The dataset provides dense coverage over thirteen predefined policies at each retained context, but it does not provide arbitrary continuous actions at every state or Bellman-consistent transitions for unrestricted offline reinforcement learning. The observed oracle is only the best candidate in this
fixed bank. A planner capable of generating a different dynamically feasible
trajectory may legitimately outperform every recorded candidate.

\textbf{Target and utility limitations.}
Binary collision, off-road, lane-departure, and unsafe-action labels were not independently validated for the final benchmark and are not used in the headline experiments. Undefined TTC values remain masked, and TTC regression was comparatively weak. The ranking utility therefore uses minimum separation,
signed displacement, jerk, maximum acceleration, and maximum yaw rate, but excludes TTC and binary safety events. Its normalization and weights encode particular trade-offs rather than a universal definition of driving safety. The favorable O-M3 policy value must be considered together with its slightly
lower mean selected separation and the failed conservative-selection audits.

\textbf{Uncertainty calibration.}
The conservative guard was calibrated on the validation partition, which had previously been used for checkpoint selection. Its empirical coverage is thus a development-set diagnostic, not an independent distribution-free or
conformal guarantee. Repeated nominal outcome replays were also unavailable, so the materiality of small action-associated outcome differences cannot be assessed relative to context-specific simulator noise.

\textbf{Test reuse.}
The same held-out partition was first used for the frozen prediction evaluation and then for an offline ranking protocol locked before ranking outcomes were accessed. This supports consistent paired comparison but does not constitute multiple independent test sets. The later conservative ensemble was therefore calibrated and analyzed on validation data and was not
evaluated as a new selector on the existing test partition. A fresh route-geometry-disjoint population is required for confirmatory execution. 

\textbf{Simulation-to-real transfer.}
All candidate outcomes are produced in CARLA. Differences in sensing, appearance, vehicle dynamics, traffic behavior, collision handling, and actor responses limit direct transfer to real vehicles. The dataset supports controlled algorithm development and simulator-based benchmarking; it does not establish real-road safety. 

\textbf{Statistical dependence and dataset scale.}
The thirteen trajectories within a branch group share a common decision context and are statistically dependent. Treating them as independent would overstate the sample size and narrow confidence intervals. All splits, bootstrap intervals, and paired tests therefore operate on complete branch groups. The principal evaluation contains 606 independent contexts, and broad generalization beyond the represented routes, towns, scenario types, seeds, and environmental configurations should not be inferred without separate held-out-domain experiments.

\subsection{Intended Use}
\label{sec:intended_use}

BranchDrive is intended for research on action-conditioned short-horizon outcome prediction, future ego-trajectory prediction, finite-bank candidate comparison, world-model validation, uncertainty estimation, and decision-model
failure analysis. The complete bank supports controlled ablations of action conditioning, history conditioning, modality choice, candidate discrimination, and offline selector value. It may also be used as pretraining or diagnostic
data for models that will subsequently undergo independent closed-loop validation.

The recommended unit of access is the complete branch group. Users should preserve the published group-disjoint split or construct new splits that keep all thirteen candidates and route-geometry variants together. Candidate policies should be represented through their recorded family, magnitude, duration, and control semantics rather than only through an integer action
index. Availability masks, replay-quality fields, provenance, and exclusion flags should be retained rather than replacing unavailable observations with apparently safe values.

The dataset is not intended to certify an autonomous vehicle, justify unsupervised real-road deployment, establish exact causal effects, or train an unrestricted offline reinforcement-learning policy. Privileged semantic BEV, actor state, expert control, and retrospective trigger information must not be
presented as onboard sensor inputs. The fixed-bank oracle must not be described as a globally optimal action, and the 7,878 trajectories must not be reported as independent statistical samples. Binary safe/unsafe labels should not be derived from TTC or separation thresholds without a separately validated
labeling and calibration protocol. 

The raw release and the strict benchmark serve different purposes. The raw collection preserves retries, partial attempts, protocol variants, and excluded groups for collection-method research. The strict 606-group release is the appropriate population for the benchmark results reported in this paper. Quality flags and manifests allow future work to define alternative subsets, but such variants should be reported separately rather than being silently combined with the frozen benchmark. 

\section{Conclusion}
\label{sec:conclusion}

This paper introduced BranchDrive, a branch-structured CARLA dataset and benchmark that associates one canonical driving history with one nominal expert future and twelve physically executed intervention futures. A read-only quality-control pipeline covering intervention compliance, temporal and modality completeness, pre-branch replay fidelity, and
future-action leakage reduced the source collection to 606 replay-valid branch groups and 7,878 associated trajectories. The resulting benchmark supports direct supervision of action-conditioned continuous outcomes and future ego trajectories while preserving the branch group as the independent statistical unit.

Models jointly conditioned on structured driving history and candidate action substantially outperformed action-only and history-only alternatives. M3 achieved scalar macro NMAE of 0.5036 and trajectory ADE of 2.2042~m, while the multimodal M5 model obtained the lowest numerical scalar error of 0.5008
without a statistically significant advantage over M3. In full-information offline evaluation, the O-M3 selector increased balanced policy value from 0.5364 to 0.5704 and reduced normalized regret from 0.2674 to 0.1495 relative to the frozen ActionPrior.

These gains do not establish safe or deployment-ready action selection. The unconstrained selector failed a predeclared minimum-separation non-worsening criterion, and a validation-calibrated global separation guard achieved high empirical coverage only by rejecting every intervention. BranchDrive
therefore validates same-context multi-action supervision for short-horizon prediction and strict fixed-bank comparison, while showing that sufficiently sharp uncertainty calibration for conservative execution remains unresolved.

\bibliographystyle{IEEEtran}
\bibliography{branchdrive_references}

\begin{IEEEbiography}
[{\includegraphics[width=1in,height=1.25in,clip,keepaspectratio]{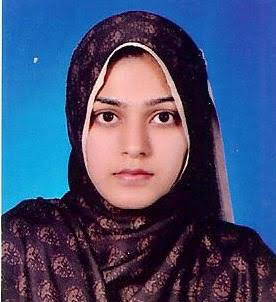}}]{Feeza Khan Khanzada} (Student Member, IEEE) received the B.E. degree in computer systems engineering from Mehran University of Engineering and Technology, Pakistan, and the M.S. degree in computer and information engineering. She is currently a graduate student research assistant and Ph.D. candidate at the University of Michigan–Dearborn. Her research interests include robotics, autonomous vehicles, deep learning for perception and control, probabilistic modeling, reinforcement learning, and robust decision-making in complex environments. She previously held research positions at Freie Universität Berlin and the University of Bath, working on machine learning, computer vision, and intelligent systems. Prior to her academic research roles, she was a software programmer with Fateh Motors Ltd., where she contributed to software development and system integration.
\end{IEEEbiography}

\begin{IEEEbiography}[{\includegraphics[width=1in,height=1.25in,clip,keepaspectratio]{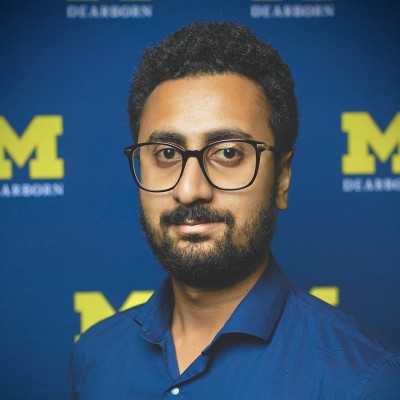}}]{Sudarshan Sridhar}
(Student Member, IEEE) is currently pursuing the M.S. degree in
computer engineering at the University of Michigan-Dearborn, Dearborn, MI, USA,
where he is a graduate student research assistant in the Intelligent Systems
Laboratory. He received the B.E. degree in
[field] from [university], [country], in [year].
His research interests include autonomous driving, counterfactual and
action-conditioned data generation, closed-loop evaluation of vision-language-action
driving models, and large-scale simulation infrastructure for high-performance
computing environments.
\end{IEEEbiography}

\begin{IEEEbiography}
[{\includegraphics[width=1in,height=1.25in,clip,keepaspectratio]{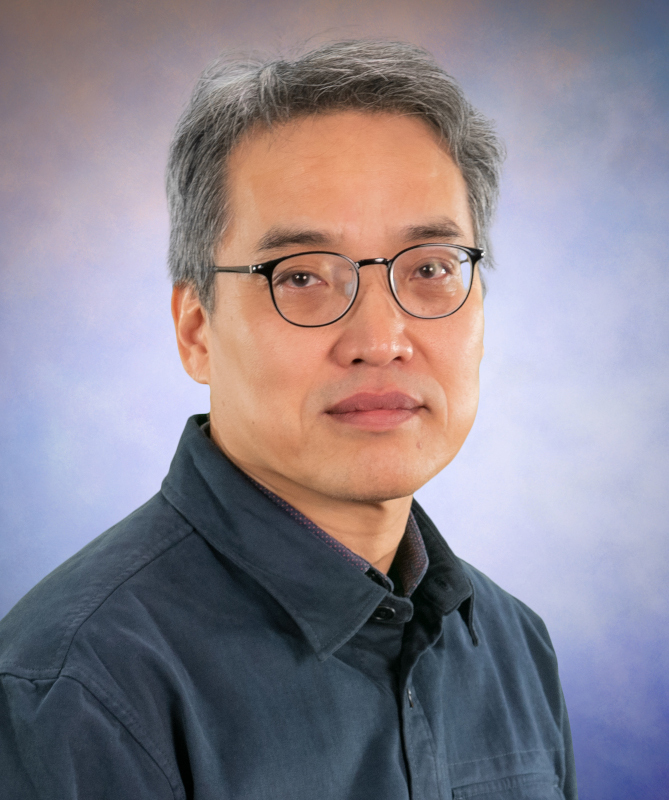}}]{Jaerock Kwon}
(Senior Member, IEEE) received the B.S. and M.S. degrees in Electronic Communication Engineering from Hanyang University, Seoul, South Korea, in 1992 and 1994, respectively, and the Ph.D. degree in Computer Engineering from Texas A\&M University, College Station, USA, in 2009. From 1994 to 2004, he worked at LG Electronics, SK Teletech, and Qualcomm Internet Services. From 2009 to 2010, he was a Professor at the Department of Electrical and Computer Engineering, Kettering University, Flint, MI, USA. Since 2010, he has been a Professor at the Department of Electrical and Computer Engineering, University of Michigan–Dearborn, MI, USA. His research interests include mobile robotics, autonomous vehicles, and artificial intelligence. His awards and honors include the Outstanding Researcher Award, the Faculty Research Fellowship (Kettering University), and the SK Excellent Employee (SK Teletech). He served as the President for the Korean Computer Scientists and Engineers Association in America (KOCSEA) in 2020, 2021, and 2025.
\end{IEEEbiography}

\end{document}